\documentclass{article}

\usepackage[preprint]{greedy_network_enlarging}

\usepackage[utf8]{inputenc} % allow utf-8 input
\usepackage[T1]{fontenc}    % use 8-bit T1 fonts
\usepackage{hyperref}       % hyperlinks
\usepackage{url}            % simple URL typesetting
\usepackage{booktabs}       % professional-quality tables
\usepackage{amsfonts}       % blackboard math symbols
\usepackage{nicefrac}       % compact symbols for 1/2, etc.
\usepackage{microtype}      % microtypography
\usepackage{xcolor}         % colors

\usepackage{amsmath, bm}
\usepackage{graphicx} 
\usepackage{subfigure}
\usepackage{wrapfig}
\usepackage{amsmath} 
\newtheorem{assumption}{Assumption}
\usepackage[ruled,vlined]{algorithm2e}
\newcommand{\eg}{\emph{e.g.}}
\newcommand{\ie}{\emph{i.e.}}
\newcommand{\etc}{\emph{etc.}}

\title{Greedy Network Enlarging}

\author{%
  Chuanjian Liu \\
  Huawei Noah's Ark Lab\\
  \texttt{liuchuanjian@huawei.com} \\
  \And
  Kai Han \\
  Huawei Noah's Ark Lab \\
  \texttt{kai.han@huawei.com} \\
  \AND
  An Xiao\\
  Huawei Noah's Ark Lab \\
  \texttt{an.xiao@huawei.com} \\
  \And
  Yiping Deng \\
  Huawei Technologies Co., Ltd. \\
  yiping.deng@huawei.com \\
  \And
  Wei Zhang \\
  Huawei Noah's Ark Lab \\
  \texttt{wz.zhang@huawei.com} \\
  \And
  Chunjing Xu \\
  Huawei Noah's Ark Lab \\
  \texttt{xuchunjing@huawei.com} \\
  \And
  Yunhe Wang \\
  Huawei Noah's Ark Lab \\
  \texttt{yunhe.wang@huawei.com} \\
}

\begin{document}

\maketitle

\begin{abstract}
	Recent studies on deep convolutional neural networks present
	a simple paradigm of architecture design, \ie, models with more MACs 
	typically achieve better accuracy, such as EfficientNet and RegNet. These 
	works try to enlarge all the stages in the model with one unified rule by 
	sampling and statistical methods. However, we observe that some network 
	architectures have similar MACs and accuracies, but their allocations on 
	computations for different stages are quite different. In this paper, we 
	propose to enlarge the capacity of CNN models by improving their width, 
	depth and resolution on stage level. Under the assumption that the 
	top-performing smaller CNNs are a proper subcomponent of the top-performing 
	larger CNNs, we propose an greedy network enlarging method based on 
	the reallocation of computations. With step-by-step modifying the 
	computations on different stages, the enlarged network will be equipped 
	with optimal allocation and utilization of MACs. On EfficientNet, our 
	method consistently outperforms the performance of the original scaling 
	method. In particular, with application of our method on GhostNet, we 
	achieve state-of-the-art 80.9\% and 84.3\% ImageNet top-1 accuracies under 
	the setting of 600M and 4.4B MACs, respectively. 
\end{abstract}

\section{Introduction} \label{section.1}
Convolutional neural networks (CNNs) deliver state-of-the-art accuracy in 
many computer vision tasks such as image classification~\cite{AlexNet,ResNet}, 
object detection~\cite{fasterRCNN}, image super-resolution~\cite{VDSR}. Most of 
deep CNNs are well designed with a predefined number of parameters and 
computational complexities. For example, ResNet~\cite{ResNet} mainly consists 
of $5$ versions with $18$, $34$, $50$, $101$ and $152$ layers. These CNNs have 
provided strong baselines for visual applications.

To improve the accuracy further, the most common way is to scale up the 
base CNN model. Three factors including depth, width and input resolution 
heavily affect the model size. A number of works propose to scale models by the 
depth~\cite{ResNet,VGGnet}, width~\cite{wide_ResNet} or
input image resolution~\cite{gpipe}. These works consider only one dimension 
from depth, width or resolution, which leads to the imbalance 
in utilization of the computations or multiply-accumulate operations (MACs) . 
Simultaneously enlarging the width, depth and resolution can provide more 
flexible design space to find the high-performance models. Recently, several 
works focus on how to efficiently scale the three factors. 
EfficientNet~\cite{efficientnet} constructed one compound scaling formula to 
constrain the network width, depth and dimension. RegNet~\cite{regnet} studied 
the relationship between width and depth by exploring the network design 
spaces. These methods utilize a unified principle to scale the whole model, but 
ignore the stage-wise differences.

\begin{wrapfigure}{r}{0.5\textwidth}
	\vspace{-1.5em}
	\begin{center}
		\includegraphics[width=0.45\textwidth]{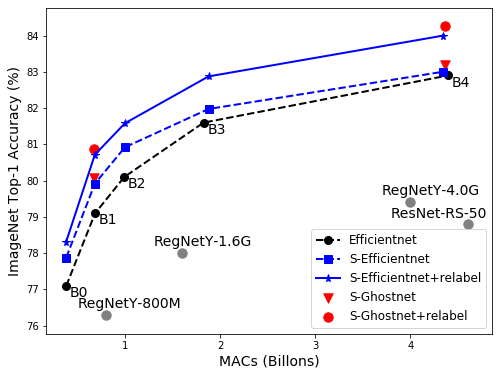}
	\end{center}
	\vspace{-1.0em}
	\caption{ImageNet classification results of our method.
		The black dash line is the original EfficientNet series. The 
		blue dash line is searched S-EfficientNet and the blue 
		solid line is S-EfficientNet with relabel trick. The red circle and 
		triangle is the performance of GhostNet-based architectures.}
	\label{fig.4}
	\vspace{-1.1em}
\end{wrapfigure}

\begin{figure}[htbp]
	\begin{center}
		\includegraphics[width=5in]{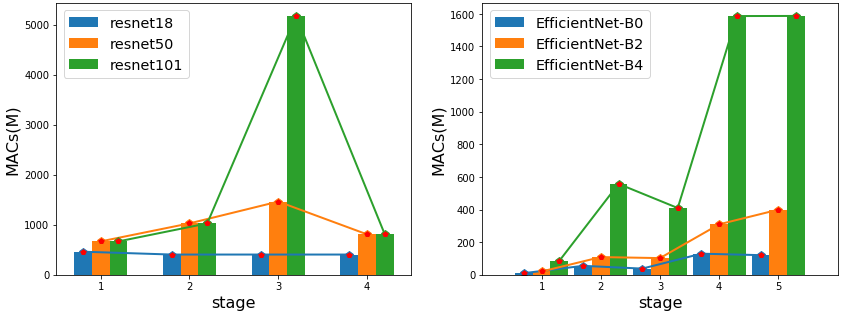}
	\end{center}
	\centering
	\vspace{-0.8em}
	\caption{MACs of different stages of CNN models. Left figure presents the 
		MACs of ResNet series. Right figure presents the MACs of 
		EfficientNet series.}
	\label{fig.1}
	\vspace{-0.3em}
\end{figure}

Here we rethink the procedure of enlarging CNN models from the viewpoint of 
stage-wise computation resource allocation. Modern CNNs usually consists of 
several stages, where one \emph{stage} contains all layers with the same 
spatial size of feature maps. In Figure~\ref{fig.1}, we present the 
computations of different stages for ResNet~\cite{ResNet} and 
EfficientNet~\cite{efficientnet}. Figure~\ref{fig.1} left demonstrate the 
discrepancy between ResNet series. ResNet18 has 
balanced MACs for each stage, while ResNet50 and ResNet101 get more MACs in the 
intermediate stages but few MACs in the head and tail stages. 
Figure~\ref{fig.1} right presents the allocation of FLOPS for EfficientNet-B0, 
EfficientNet-B2 and EfficientNet-B4. EfficientNet utilizes one unified model 
scaling principle for network width, depth and resolution, so different 
configurations of EfficientNet have the similar tendency of MACs on different 
stages. The later stages have far more MACs (\textgreater $10$ times) than the 
former and intermediate stages. A universal rule of the computation allocation 
for different models is impractical. Neither the manual designed or unified 
rule is the solution of optimal computations allocation.

%Besides, many structured network pruning method\cite{liu2019learning,netslim} 
%got different prune ratio for different layers. Generally, the lower layers in 
%CNN usually learn the local patch information. The higher layer contains more
%global semantic information. As a result, different stages tend to have 
%different demands on computations. Usually, the pruned networks have less 
%computation redundancy. However, very large initial network are needed for the 
%top-down prune method. In this paper, we achieve the purpose of optimal 
%allocation of computations by designing one bottom-up search method.

In this paper, we propose a network enlarging method based on greedy search of 
computations for each stage. In contrast to conventional unified principle, the 
method performs fine-grained search on the reallocation of computations. Given 
a baseline network, our goal is to enlarge it to the target MACs with the best 
configuration of depth, width and resolution in each stage. Under the 
assumption that the top-performing smaller CNNs are a proper subcomponent of 
the top-performing larger CNNs, we are able to enlarge CNNs step-by-step using 
greedy network enlarging algorithm. For each iteration in proposed algorithm, 
1) a series of candidate networks are constructed by searching width, depth and 
resolution of each stage under constrained MACs; 2) with fast performance 
evaluation method, the architecture with the best performance in this iteration 
is appended to the baseline model pool for next iteration. By gradually adding 
MACs at each iteration, we find the optimal architecture until achieving the 
target MACs. Experiment results on ImageNet classification task demonstrate the 
superiority of our proposed method. The searched network configurations can 
largely boost the performances of existing base models. For example, searched 
EfficientNet models by proposed method outperform the original 
EfficientNets by a large margin. 

%In summary, our main contributions are as follow:
%\begin{itemize}
%	\item[1)] 
%	We propose one greedy enlarging model method to acquire the optimal network 
%	architecture which maximize the utilization of MACs.    
%	\item[2)] 
%	We did experimental analysis on the fast performance evaluation method and 
%	select the most relevant method with function-preserving transformation.
%	\item[3)]
%	Compared with the conventional enlarging model method, the 
%	proposed greedy method can provide fine grained network architecture 
%	configuration. Better combinations of width, depth and resolution improved 
%	the utilization of MACs greatly. 
%	\item[4)]
%	We demonstrate the effectiveness of greedy enlarging model method on 
%	several state-of-art networks. With a similar computation budget, we can 
%	improve $1.35\%$ accuracy on ImageNet.
%\end{itemize}

\section{Related Work}
\paragraph{Manual Network Design.} In the early days after 
AlexNet~\cite{AlexNet}, a large number of manually designed network 
architecture emerged. VGG~\cite{VGGnet} is the typical CNN architecture without 
any special connections, and deeper VGG-nets get high accuracies. However, the 
convergence problem emerged for very deep network. ResNet~\cite{ResNet} with 
shortcut was proposed with higher accuracy and more layers. Except deeper 
network, wider network is another direction. 
WideResnet~\cite{wide_ResNet} has higher accuracy by adding channels for each 
layer in Resnet. Besides, a number of light-weight network are proposed in 
order to meet the demands of mobile devices. GoogLeNet~\cite{GoogleNet}, 
MobileNets~\cite{Mobilenet, Mobilenet2}, 
ShuffleNets~\cite{Shufflenet,shufflenetv2} and GhostNet~\cite{ghostnet}are 
these type networks. By setting one width scaling factor, the accuracy and MACs 
of Mobilenets and GhostNet are improved. The design 
pattern behind these networks was largely man-powered and focused on 
discovering new design choices that improve accuracy, \eg, the use of deeper or 
wider models or shortcuts. 

\paragraph{Automatic Network Design.} Currently expert designed architectures 
are time-consuming. Because of this, there is a growing interest in automated 
neural architecture search (NAS) methods~\cite{elsken2019neural}. By now, NAS 
methods have 
outperformed manually designed architectures on some tasks such as image 
classification~\cite{nas1,nas2,nas3,darts,Mobilenetv3}, object
detection~\cite{nas_fpn,auto_fpn,spnas,efficientdet} or semantic 
segmentation~\cite{auto_deeplab,squeezenas}. Generally, more MACs 
means higher accuracy. Traditionally, researchers have already learned to 
change the depth, width or resolution of models. But only one dimension is 
considered usually. EfficientNet~\cite{efficientnet} showed that it was 
critical to balance all dimensions of network width/depth/resolution and 
proposed a simple yet effective compound scaling method in accordance with the 
results by random sampling. RegNet~\cite{regnet} got several patterns by a huge 
number of experiments: $1.$ good network have increasing widths with stages; 
$2.$ the stage depths are likewise tend to increase for the best models, 
although not necessarily in the last stage. These methods construct principles 
from small networks, and use the rule to get various sizes of model, even very 
large models. In this paper, we take use of greedy allocation of MACs to 
enlarge model and get the specific model architecture under constrained MACs. 
During the expansion, the width, depth and input resolution are considered for 
each stage. Our intention is to maximize the utilization of MACs for the 
network.

\section{Approach}
In this section, we describe the proposed network enlarging method based on 
greedy allocation of MACs. Firstly, we define the goal of our method to find 
the optimal depth and width of each stage and the input resolution. Secondly, 
we introduce  
the main algorithm of greedy network enlarging. Further, we introduce how to 
efficiently evaluate the performance of candidate models.

\begin{figure*}[t]
	\vspace{-1.0em}
	\begin{center}
		\includegraphics[width=5in]{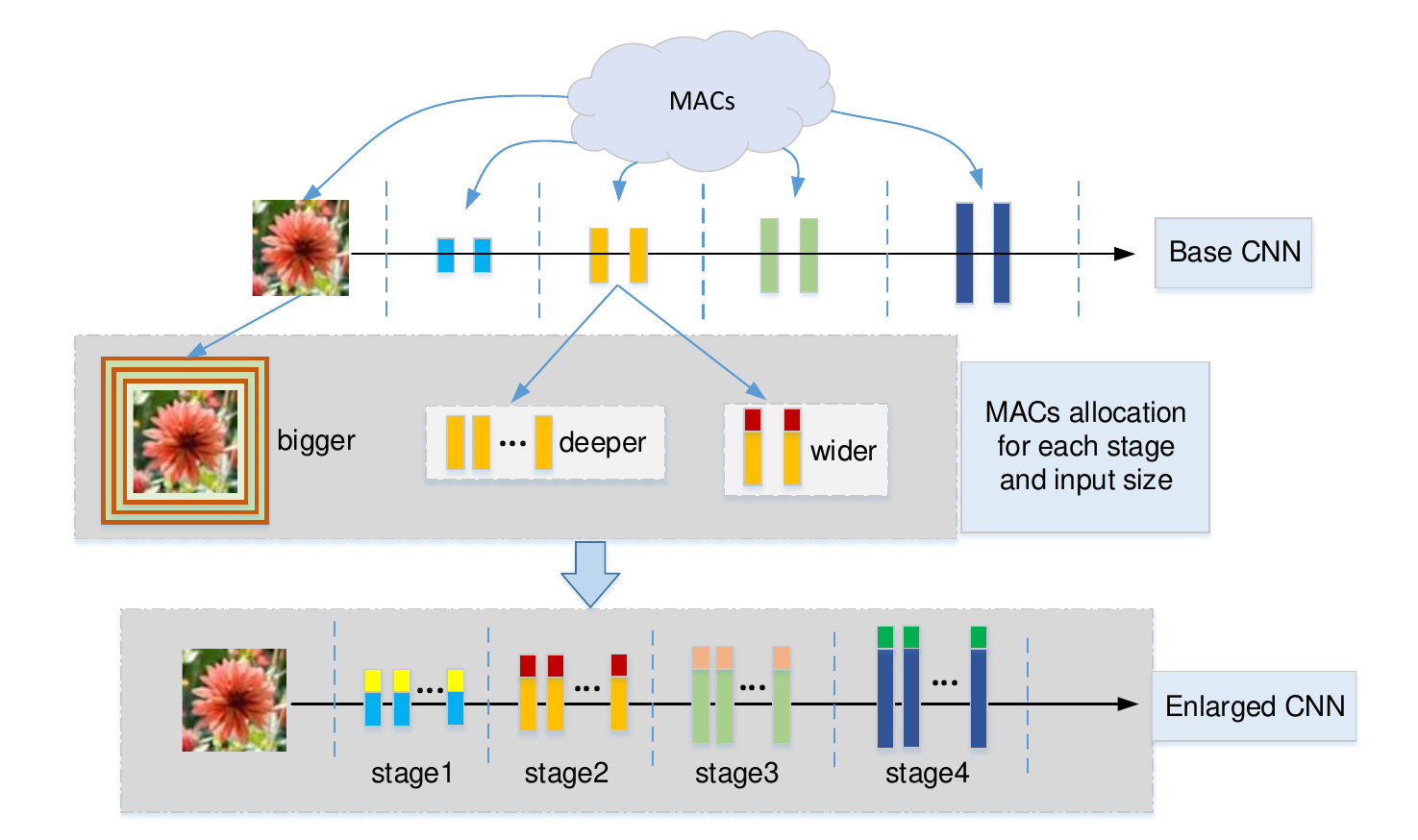}
	\end{center}
	\vspace{-0.5em}
	\caption{The framework for adjusting input resolution, width and depth for each stage. The surrounding box out of the input image means candidate input resolution.}
	\label{fig.2}
	\vspace{-1em}
\end{figure*}

\subsection{Problem Definition}
The modern CNN backbone architectures usually consists of a stem layer, network 
body and a head~\cite{regnet,ResNet,efficientnet}. The main MACs and 
parameters burdens lie in the network body, as typically the stem layer is a 
convolutional layer and the head is a fully-connected layer. Thus, in this 
paper we focus on the scaling strategy of network body. The network body 
consists of several \emph{stages}~\cite{regnet}, which are defined as a 
sequence of layers or blocks with the same spatial size. For example, 
ResNet50~\cite{ResNet} body is composed of $4$ stages with $56\times56$, 
$28\times28$, $14\times14$, and $7\times7$ output sizes, respectively.

Scaling up convolutional neural networks is widely used to achieve better 
accuracy. Network depth, width and input resolution are three key factors for 
model scaling. Deeper convolutional neural networks capture richer and more 
complex features, and usually have high performance in contrast to shallow 
network. With the help of shortcuts, very deep network can be trained to 
convergence. However, the improvements on accuracy become smaller with the 
increase of depth. Another direction is scaling the width of network. More 
kernels mean more fine grained features can be learned. However, the MACs is 
squared with the width. As a result, the network depth will be constrained and 
high level features maybe loss. EfficientNet~\cite{efficientnet} showed that 
the accuracy quickly saturates when networks become wider. Higher resolution 
provide rich fine-grained information. In order to match high resolution, more 
powerful network is wanted. Deeper and wider network can acquire large 
receptive field and capture fine grained features. 

As a result, the network depth, width and resolution are not independent. These 
three dimension have various combinations. And one unified principle can not 
acquire the best configuration for all tasks. In this paper, we decompose the 
network depth, width and resolution into stage depth, width and input 
resolution. This will maximize the utilization of computations for each 
stage and the whole network.

Given a base network $\mathbb{N}$ with $L$ stages, width and depth are 
$\mathbf{w,d}=(w_{1}, w_{2}, ..., w _{L}, d_{1}, d_{2}, ..., d _{L})$, and 
input resolution is $\mathbf{r}$. The objective is to acquire the network 
architecture with best performance by optimizing the allocation of target 
MACs $T$:
\begin{align}
	\label{defination}
	\mathbf{r^\star,d^\star,w^\star} &= 
	\arg\mathop{\max}\limits_{\mathbf{r,d,w}}\mathrm{ACC}_{\mathrm{val}}(\mathbb{N}(\mathbf{r,d,w},\mathbf{\theta^\star}))
	\\
	\mathrm{s.t.} &\quad\ |MACs(\mathbb{N}(\mathbf{r,d,w}))-T| \leq 
	{\delta\cdot T} 
	\\
	\mathbf{\theta^\star} &= 
	\arg\mathop{\min}_{\mathbf{\theta}}\mathrm{LOSS}_\mathrm{train}(\mathbb{N}(\mathbf{r,d,w},\mathbf{\theta}))
\end{align}
where $\mathbf{\theta}$ is the trainable parameters of the network. 
$\mathrm{ACC}_\mathrm{val}$ denote the validation accuracy, $T$ is the target 
threshold of MACs and $\delta$ is used to control the difference between the 
MACs of the searched model and the target MACs.

\textbf{Search space.} We consider the combinations of input resolution, width 
and depth of each stage. Suppose a base network with $L$ stages and 
configurations of width and depth $\mathbf{w,d}$ 
=  ($w_{1}$, $w_{2}$, ..., $w _{L}$, $d_{1}$, $d_{2}$, ..., $d _{L}$), and 
input resolution $\mathbf{r}$. For each tieration, the growth rate for width, 
depth and resolution is $s_w$, $s_d$ and $s_r$, respectively. Under constrained 
target MACs, we enlarge the width, depth for each stage and the network 
input resolution step-by-step. For example, ResNet18 contains $4$ stages, if we 
constrain the search upper bound is $3$ times in both depth and width for each 
stage, and the growth rate is $1$ for depth and width. The total number of 
combinations is 
$128\times256\times512\times1024\times4\times4\times4\times4\approx 
4.4\times10^{12}$ without considering the variation of input size.

%-------------------------------------------------------------------------
\subsection{Algorithm of Greedy Network Enlarging}
Figure~\ref{fig.2} presents our framework. Our intention is to find the optimal 
allocation of computations by enlarging depth, width and input resolution for 
each stage under constrained computations. So as to maximize the utilization of 
MACs, as shown in Eq.~\ref{defination}. For each stage $i$ in the network, we 
try to find its optimal depth $d_i^{\star}$ and width $w_i^{\star}$. The 
optimal input size $r^{\star}$ of resolution is searched to match the 
specialized width and depth. In the problem\ref{defination}, $\mathbf{r,d,w}$ 
have discrete values and massive combinations.  Deep learning is both time and 
resource consuming. Due to the extreme complexity, traditional mathematical 
optimization method is impracticable. So we turn to efficient neural 
architecture search method.

To simply the search complexity, we first introduce an assumption. Finding the 
global optimal model is difficult with the massive search space, so we can 
smooth the target to find a top-performing configuration of target MACs. 
%Since the ways to set the depth, width and resolution to obtain the target 
%%%MACs are numerous, there are many the opportunities to achieve top 
%%%%performance. 
We introduce an assumption that the top-performing smaller CNNs are a 
proper subcomponent of the top-performing larger CNNs, as shown in 
Assumption~\ref{as:1}. Resnet~\cite{ResNet}, VGG~\cite{VGGnet}, 
EfficientNet~\cite{efficientnet} \etc, fit 
this assumption perfectly. This assumption enables the idea of efficient search 
algorithm via greedy network enlarging.

\begin{assumption}\label{as:1}
	Given an optimal network with MACs $B_0$, depth 
	$(d^0_1,d^0_2,\cdots,d^0_L)$, width $(w^0_1,w^0_2,\cdots,w^0_L)$ and 
	resolution $r^0$, there exists at least one top-performing network with 
	MACs $B_0+\delta$, depth $(d^1_1,d^1_2,\cdots,d^1_L)$, width 
	$(w^1_1,w^1_2,\cdots,w^1_L)$ and resolution $r^1$ that satisfies
	\begin{equation}
	\begin{aligned}
	d^0_i &\leq d^1_i,~\forall i=1,2,\cdots,L;\\
	w^0_i &\leq w^1_i,~\forall i=1,2,\cdots,L;\\
	r^0 &\leq r^1.
	\end{aligned}
	\end{equation}
\end{assumption}

With the above assumption, we transform the optimal network architecture search 
problem into a series of interrelated single-stage optimal sub-network 
architecture search problems, and then solve them one by one. Decisions need to 
be made at each stage to optimize the process. The selection of decisions at 
each stage depends only on the current state (here, the current state refers to 
the resolutions, widths and depths of the current stage). When the decision of 
each stage is determined, a decision sequence is formed, which determines the 
final solution. The overall algorithm is illustrated in 
Algorithm~\ref{algorithm.1}.

%In computer science, if a problem can be solved optimally by breaking it into 
%sub-problems and then recursively finding the optimal solutions to the 
%sub-problems, then it is said to have optimal substructure. To this end, we 
%utilize dynamic programming algorithm to optimize the search progress. In 
%this 
%procedure, the {current state} is defined as one set of optimal 
%configurations 
%searched before this iteration. The overall algorithm is illustrated in 
%Algorithm~\ref{algorithm.1}.

\begin{algorithm}[htbp]
	\SetAlgoLined
	\label{algorithm.1}
	\KwResult{Configurations 
	$\mathbf{c^{\star}}=\mathbf{r^{\star},d^{\star},w^{\star}}$ with 
		target MACs $\mathbf{T}$.}
	\textbf{Initialization:} Base network $\mathbb{N}$ with $B_0$ MACs and 
	$\mathbf{L}$ 
	stages: width 
	$\mathbf{w^0}$ 
	=  ($w^0_{1}$, $w^0_{2}$, ..., $w^0_{L}$), depth $\mathbf{d^0}$ 
	=  ($d^0_{1}$, $d^0_{2}$, ..., $d^0_{L}$) and input resolution 
	$\mathbf{r^0}$. Total dimension of search space is $(2L+1)$. The target 
	MACs of output network is $\mathbf{T}$ and the rate of error is $\delta$. 
	The search number is $\mathbf{N}$. Initialize the set of optimal 
	sub-configurations as 
	$\bm{S}=\{(\mathbf{r^{0}},\mathbf{d^{0}},\mathbf{w^{0}})\}$, the growth 
	rate of resolution $s_r$\;
	\While{$i\leq\mathbf{N}$}{
		current target MACs: 
		$T_i=B_0*(\frac{\mathbf{T}}{B_0})^{\frac{i}{\mathbf{N}}}$;
		
		current candidates $\bm{C}=[\ ]$;
		
		\For {$(\mathbf{r},\mathbf{d},\mathbf{w})$ in $\bm{S}$}{
			
			\While{$MACs(\mathbb{N}(\mathbf{r},\mathbf{d},\mathbf{w}))<T_i$}{
				$\mathbf{r} = \mathbf{r}+s_r$\;
			}
			\If{$|MACs(\mathbb{N}(\mathbf{r},\mathbf{d},\mathbf{w}))-T_i|\leq 
				\delta\cdot T$}{$\bm{C}$ append 
				$(\mathbf{r},\mathbf{d},\mathbf{w})$}
		}
		\For {$(\mathbf{r},\mathbf{d},\mathbf{w})$ in $\bm{S}$}{
			\For {$j$ in $\mathit{range}(L)$}{
				\While{$MACs(\mathbb{N}(\mathbf{r},\mathbf{d},\mathbf{w}))<T_i$}{
					$\bm{C}' = proportional\ collection(\mathbf{d},\mathbf{w})$ 
					as in 
					Algorithm~\ref{algorithm.2};
				}
				$\bm{C}$ extend 
				$\bm{C}'$
				%				
				%\If{$|MACs(\mathbb{N}(\mathbf{r},\mathbf{d},\mathbf{w}))-T_i|\leq
				%					\delta\cdot T$}{$\bm{C}$ append 
				%					$(\mathbf{r},\mathbf{d},\mathbf{w})$}
			}
			
		}
		$\mathit{ACC} = [\ ]$\;
		\For {$\mathit{c}$ in $\bm{C}$}{
			$acc = performance\ evaluation(\mathit{c})$ as stated in 
			Sec.~\ref{sec:performance}\;
			$\mathit{ACC}$ append $acc$\;
		}
		$\mathit{index} = \arg\mathop{\max}\mathit{ACC}$\;
		$\bm{S}$ append $\bm{C_{index}}$\;
		\If{$i==N$}{
			$\mathbf{c^{\star}} = \bm{C_{index}}$\;
		}
	}
	\caption{Greedy network enlarging.}
\end{algorithm}

In the algorithm, we use exponential increment of MACs in the process of 
search. This way make the changes of network more gentaly in contrast to 
uniform increment. For each iteration in Algorithm~\ref{algorithm.1}, in order 
to find the local optimal architecture configuration, we have to search and 
evaluate the candidate architectures. This step contains two targets: the first 
is to find the candidate architectures under limited increase of MACs; the 
second is to find the local optimal architectures with maximum validation 
accuracy. In the step of acquiring candidates, we consider the increase of 
resolution separately, which reduces the candidates. The increase of width and 
depth of each stage is on the basis of corresponding resolution.

\begin{algorithm}[htbp]
	\SetAlgoLined
	\label{algorithm.2}
	\KwResult{Candidates $\bm{C}$ with collected width $w$ and depth $d$}
	\textbf{Initialization:} Target MACs $T_i$, configuration 
	$(\mathbf{r},\mathbf{d},\mathbf{w})$, ratios set $\bm{P}$. The growth rate 
	of depth and width is $s_d$ and $s_w$, respectively, current stage $j$\;
		$d_j\in\mathbf{d}$;	$w_j\in\mathbf{w}$\;
		\For{$p\in\bm{P}$}{
			$T_d = p*T_i$\;
			\While{$MACs(\mathbb{N}(\mathbf{r},\mathbf{d},\mathbf{w}))\le 
			T_d$}{
				$d_j = d_j + s_d$
			}
			\While{$MACs(\mathbb{N}(\mathbf{r},\mathbf{d},\mathbf{w}))\le 
			T_i$}{
				$w_j = w_j + s_w$
			}
		\If{$|MACs(\mathbb{N}(\mathbf{r},\mathbf{d},\mathbf{w}))-T_i|\leq 
			\delta\cdot T$}{$\bm{C}$ append 
			$(\mathbf{r},\mathbf{d},\mathbf{w})$}
		}
	\caption{Proportional collection of width and depth.}
\end{algorithm}

In order to reduce the searched candidates, we take use of proportional control 
factor to assign the MACs between depth and width for each stage. Specifically, 
the ratios of MACs between depth and width are in one set $\bm{P}$. 
Under this setting, we search depth first and then width for each stage. The  
algorithm is illustrated in Algorithm~\ref{algorithm.2}.

\subsection{Performance Estimation}\label{sec:performance}
To guide the search process, we have to estimate the performance of a given 
architecture. The most accurate method is to train the candidates on the 
whole training data and evaluate their performance on validation data. However, 
this way requires great computational demands in the order of thousands of GPU 
days. Developing methods for speeding up the process of performance estimation 
is crucial.

We turn to proxy tasks to estimate performance. Including 
shorter training times~\cite{regnet}, training on a 
subset of the data~\cite{pmlr-v54-klein17a}, on proxy data~\cite{nas1} or 
using less filters per layer and less cells~\cite{Real_Aggarwal_Huang_Le_2019}. 
These low-fidelity approximations reduce the cost, they 
also introduce bias in the performance estimation. Proxy data and simplified 
architecture have large deviation which leads to poor rank preservation. 

In this section, we determine the optimal proxy task for performance estimation 
with empirical experiments. Firstly, we get the proxy sub-dataset by evaluating 
the performance of different sub-datasets. Secondly, the hyper-parameters of 
training are acquired with parameter search. Spearman's rank correlation 
coefficient $\rho$ is a non-parametric measure of rank correlation, which is 
used as the measure of proxy task.

For the proxy sub-dataset, we create two sub-datasets ImageNet1000-100 and 
ImageNet100-500 by random selecting images from ImageNet. To evaluate these 
datasets, $12$ 
network architectures with different width, depth and input sizes are generated 
on the basis of EfficientNet-B0. We train all the $12$ networks and 
EfficientNet-B0 on the whole train set of ImageNet for $150$ epochs, the Top-1 
accuracies on the validation dataset are used as the comparison object. We 
finetune the $12$ networks for different epochs. Besides, we 
train the 
$12$ networks from scratch for few epochs. On ImageNet100-500, the average 
Spearman value is $\rho=0.16$. On 
ImageNet1000-100, the average Spearman value is $\rho=0.23$. So we choose 
ImageNet1000-100 as the proxy sub-dataset. More details are presented in the 
supplementary materials.

After determining the proxy dataset, we try to improve the correlation between 
the proxy task and original task by searching the hyper-parameters. $24$ 
network architectures with different width, depth and input sizes are generated 
on the basis of EfficientNet-B0. We train all of the $24$ networks on the whole 
train set of ImageNet for $150$ epochs, the Top-1 accuracies on the validation 
dataset are used as the comparison object. Two pretrained EfficientNet-B0 
models on the ImageNet and ImageNet1000-100 are provided, respectively. 
The learning rate, mode of learning rate decay and training 
epochs are considered. Among these hyper-parameter combinations, 
the top-2 Spearman value $\rho$ is $0.57$ and $0.54$, these values indicate 
moderate positive correlation. They both use cosine decay method and the 
initial learning rate is $0.01$ for training $20$ epochs. The difference is 
that the first use the ImageNet1000-100 pretrained model and the second use the 
ImageNet pretrained model. More details are presented in the supplementary 
materials. Figure.~\ref{fig.3} presents the consistency of different networks. 
In the next section, we take use of initial learning rate is $0.01$ and cosine 
decay for finetuning $20$ epochs on the ImageNet1000-100 pretrained model.

\begin{figure*}[htbp]
	\begin{center}
		\includegraphics[width=4in]{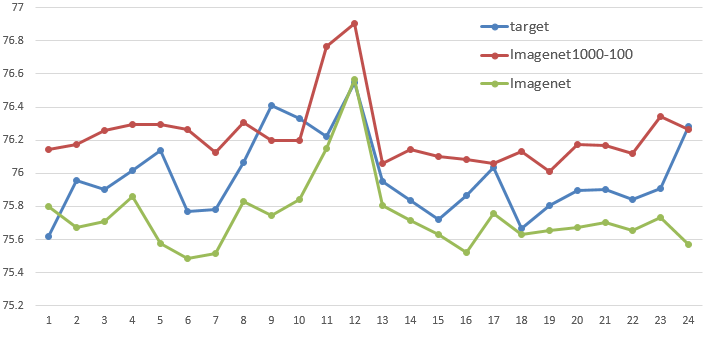}
	\end{center}
	\vspace{-0.5em}
	\caption{Correlation between the proxy task and original task. The blue line is the target. The red line has Spearman value 
	$0.57$, it is trained on the basis of ImageNet1000-100. The cyan line has 
	spearman value $0.54$ which is trained on the basis of ImageNet. For 
	comparison purposes, we manually adjust the accuracy up by $27$ and $10$ 
	for red line and cyan line, respectively.}
	\label{fig.3}
	\vspace{-1.0em}
\end{figure*}

\section{Experiments}\label{experiments}

In this section, we evaluate greedy network enlarging method on general 
image classification dataset ImageNet~\cite{imagenet}. We demonstrate the 
method gets state-of-the-art accuracy with similar MACs.

\subsection{Datasets, Networks and Experimental Settings}

We extensively evaluate our methods on popular classification datasets 
ImageNet(ILSVRC2012)~\cite{imagenet}, which contains $1.3$M images and $1000$ 
categories, the validation set contains $50$K images. On ImageNet, in order to 
speed up the search process, we create proxy ImageNet1000-100 dataset, which 
contains $100$K train images and $50$K validation images randomly sampled from 
ImageNet train set. Two baseline networks are considered: 
EfficientNet~\cite{efficientnet} and improved 
GhostNet\url{https://gitee.com/mindspore/models/tree/master/research/cv/ghostnet}~\cite{ghostnet}.

To accelerate the search process, we set the growth rate of resolution and 
depth as $s_r=8$ and $s_d=1$, respectively. For the growth rate of width, we 
use $s_w=2$ for small model and $s_w=4$ for large model. The ratios of MACs 
between depth and width are in one set $\bm{P}=\{0.0,0.1,0.2,\cdots,1.0\}$ The 
error rate of MACs 
is $\delta=0.01$. We take use of exponential growth of MACs. We set different 
number of search iterations $\mathbf{N}$ for small and large models. The 
finetune method 
comes from function preserving algorithm~\cite{chen2015net2net}.

After the process of search is completed, we retrain the acquired network 
architecture on the whole ImageNet from scratch. The train setting is from 
timm~\cite{rw2019timm} under its license and EfficientNet~\cite{efficientnet}. 
RMSProp optimizer 
with momentum 0.9; weight decay 1e-5; 
multi-step learning rate with warmup, initial learning rate 0.064 that decays 
by 0.97 every 2.4 epochs. Moving average of weight, 
dropblock~\cite{ghiasi2018dropblock}, random erasing~\cite{zhong2020random} and 
random augment~\cite{cubuk2020randaugment} are used. 

ImageNet has noise labels and the method of crop augmentation introduces more 
noisy input and labels. To prevent this, we use the relabel 
method~\cite{yun2021re} to get higher accuracy. 

\subsection{ImageNet Results and Analysis}

For EfficientNet, we take EfficientNet-B0 as the baseline, and we search the 
models with MACs similar to EfficientNet-B$1$ to B$4$. Besides, we enlarge 
GhostNet with the principle of EfficientNet and search 
GhostNet architectures with greedy search method. For GhostNet, we add 
Squeeze-and-Excitation~\cite{senet} 
module for each block. Table.\ref{tab.1} shows the main results and comparison 
with other networks. The searched models are marked with '\textbf{S-}'.

GhostNet-B1 and GhostNet-B4 in~\ref{tab.1} are obtained by the 
compounding scale rule of EfficientNet. Their performance is lower in contrast 
to greedy search methods. This suggests that the rule on EfficientNet is not 
fit for GhostNet. We need to resample and optimize for new networks to get 
suitable rules. Besides, the compounding scale principle ignores the difference 
of stages, which leads to the loss of elaborate adjustment.

\begin{table*}[htbp]
	\centering
	\tiny
	\resizebox{1.0\textwidth}{!}{
		\begin{tabular}{l|cc|cc}
			\toprule [0.15em]
			Model &  Top-1 Acc. & \#Params &\#MACs & Ratio-to-EfficientNet  \\
			\midrule [0.1em]
			EfficientNet-B0~\cite{efficientnet}  & 77.1\% & 5.3M  & 0.39B &  1x 
			\\
			Ghostnet $1.0\times$~\cite{ghostnet} & 73.9\% & 5.2M & 0.14B & 
			0.36x \\
			
			\midrule
			EfficientNet-B1~\cite{efficientnet}  & 79.1\% & 7.8M  & 0.69B &  1x 
			\\
			ResNet-RS-50~\cite{bello2021revisiting} & 78.8\% & 36M & 4.6B & 6.7x\\
			REGNETY-800MF~\cite{regnet} & 76.3\% & 6.3M & 0.8B & 1.16x \\
			\bf S-EfficientNet-B1 & \bf 79.91\% & 8.8M & 0.68B & 1x \\
			\bf S-EfficientNet-B1-re & \bf 80.71\% & 8.8M & 0.68B & 1x \\
			GhostNet-B1 & 79.13\% & 13.3M & 0.59B & 0.85x \\
			\bf S-GhostNet-B1 & \bf 80.08\% & 16.2M & 0.67B & 1x \\
			\bf S-GhostNet-B1-re & \bf 80.87\% & 16.2M & 0.67B & 1x \\
			
			\midrule
			EfficientNet-B2~\cite{efficientnet}  & 80.1\% & 9.1M  & 0.99B &  1x 
			\\
			REGNETY-1.6GF~\cite{regnet} & 78.0\% & 11.2M & 1.6B & 1.6x \\
			\bf S-EfficientNet-B2 & \bf 80.92\% & 9.3M & 1.0B & 1x \\
			\bf S-EfficientNet-B2-re & \bf 81.58\% & 9.3M & 1.0B & 1x \\
			
			\midrule
			EfficientNet-B3~\cite{efficientnet}  & 81.6\% & 12.2M  & 1.83B &  
			1x \\
			ResNet-RS-101~\cite{bello2021revisiting} & 81.2\% & 64M & 12B & 6.6x\\
			REGNETY-4.0GF~\cite{regnet} & 79.4\% & 20.6M & 4.0B & 2.18x \\
			\bf S-EfficientNet-B3 & \bf 81.98\% & 12.3M & 1.88B & 1x \\
			\bf S-EfficientNet-B3-re & \bf 82.87\%  & 12.3M & 1.88B & 1x \\
			
			\midrule
			EfficientNet-B4~\cite{efficientnet}  & 82.9\% & 19.3M  & 4.39B &  
			1x \\
			REGNETY-8.0GF~\cite{regnet} & 79.9\% & 39.2M & 8.0B & 1.82x \\
			NFNet-F0~\cite{brock2021high} & 83.6\% & 71.5M & 12.4B & 2.8x \\
			ResNet-RS-152~\cite{bello2021revisiting} & 83.0\% & 87M & 31B & 7.1x \\
			EfficientNetV2-S~\cite{tan2021efficientnetv2} & 83.9\% & 24M & 8.8B 
			& 2.0x \\
			\bf S-EfficientNet-B4 & \bf 83.0\% & 17.0M & 4.34B & 1x \\
			\bf S-EfficientNet-B4-re & \bf 84.0\% & 17.0M & 4.34B & 1x \\
			GhostNet-B4 & 82.78\% & 36.1M & 4.39B & 1x \\ 
			\bf S-GhostNet-B4 & \bf 83.2\% & 32.9M & 4.37B & 1x \\
			\bf S-GhostNet-B4-re & \bf 84.3\% & 32.9M & 4.37B & 1x \\
			
			\bottomrule[0.15em]
		\end{tabular}
	}
	\caption{
		{Searched Architecture Performance on ImageNet}. The '-re' means 
		the models are trained with relabel trick~\cite{yun2021re}. Our results are in \textbf{bold}.
	}
	\label{tab.1}
	\vspace{-1.0em}
\end{table*}

In Table.\ref{tab.1}, Top-1 accuracies of all searched architectures outperform 
the compound scaling tricks of EfficientNet~\cite{efficientnet} and 
RegNet~\cite{regnet}. On $600$M MACs, our searched architectures get $79.91\%$ 
and $80.08\%$, improve performance $0.81\%$ and $0.97\%$, respectively. On 
EfficientNet-B2 and B3, our searched EfficientNet architectures achieve 
$80.92\%$ and $81.98\%$. We search $2$ networks on $4$B MACs level, 
S-EfficientNet-B4 gets $83.002\%$ and S-GhostNet-B4 gets $83.2\%$, 
respectively. 

The relabel training trick improve the accuracy further. The Top-1 accuracy 
improves $0.6\%$ to $1.1\%$ on all searched architectures. We achieve a new 
SOTA 80.87\% and 84.3\% ImageNet top-1 accuracy under the setting of $600$M and 
$4.4$B MACs, respectively. All searched network architectures are presented in 
the supplementary materials.

\subsection{Process of Greedy Search}

\begin{figure*}[htbp]
	\vspace{-1.0em}
	\begin{center}
		\includegraphics[width=5in]{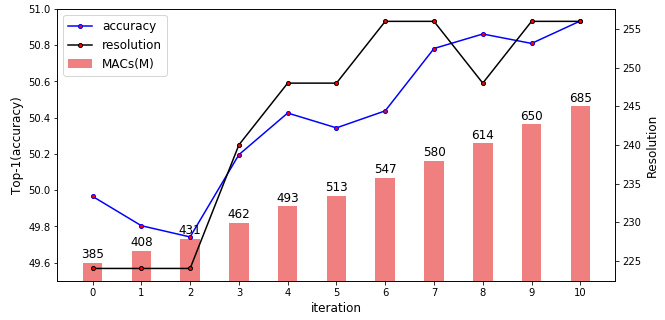}
	\end{center}
	\vspace{-0.5em}
	\caption{The search process of target $686$M MACs of EfficientNet-B1. The 
		blue and black line demonstrate the changes of accuracy and input size 
		as the increase of MACS, respectively.}
	\label{fig.5}
	\vspace{-1.0em}
\end{figure*}

Figure~\ref{fig.5} is used specifically to show the changes of accuracy and 
input resolution of the search process. With increase of MACs, the resolution 
rises wavily, which verifies the role of dynamic search. The accuracy increases 
slowly and steadily.

\begin{figure*}[htbp]
	\vspace{-1.0em}
	\begin{center}
		\includegraphics[width=5in]{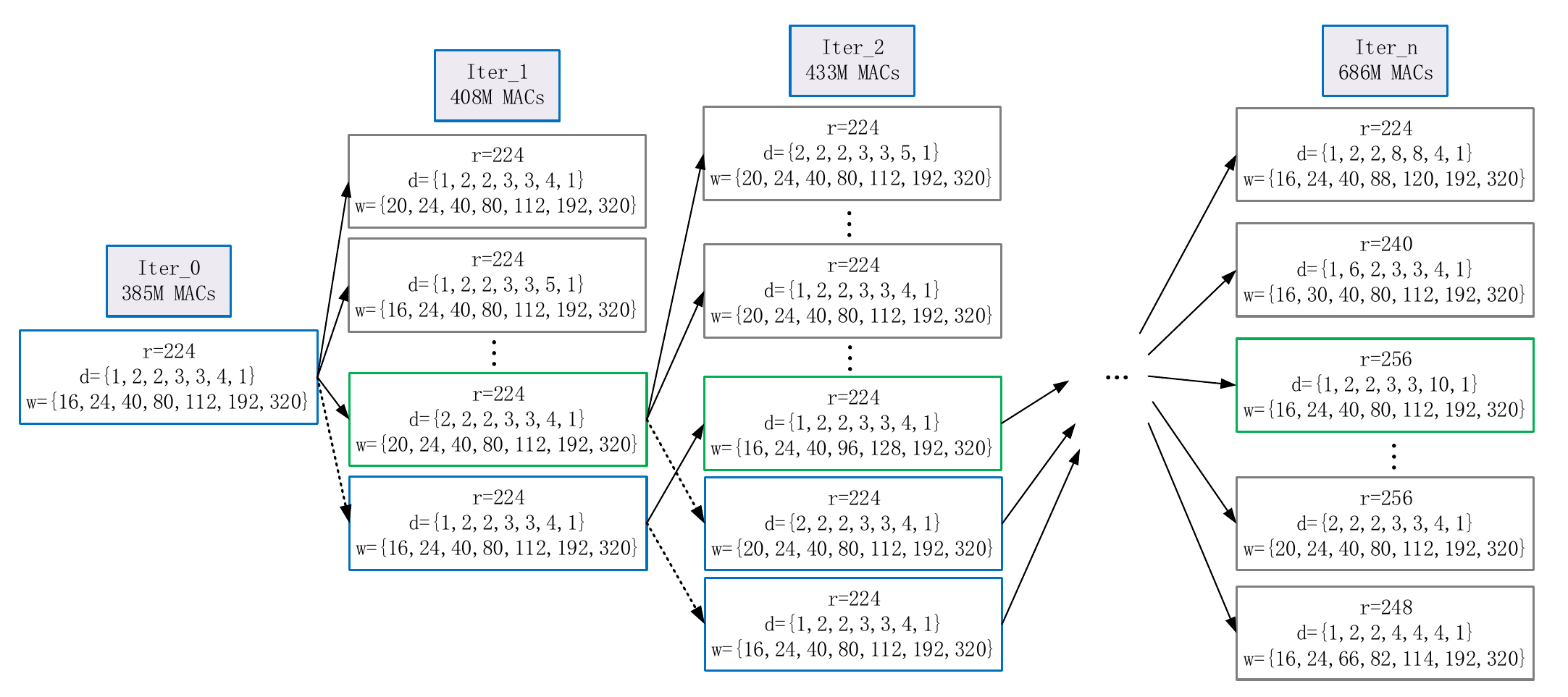}
	\end{center}
	\vspace{-0.5em}
	\caption{The schematic diagram of greedy search for EfficientNet-B1.}
	\label{fig.6}
	\vspace{-1.0em}
\end{figure*}

Furtherly, the schematic diagram of greedy search for EfficientNet-B1 is shown 
in Figure~\ref*{fig.6}. Under constrained MACs, we show the candidate network 
architectures. The green box means the best architecture in current iteration, 
and the gray box are discarded. Besides, the best architecture of each 
iteration are delivered to the later iterations.

\section{Conclusion} \label{conclusion}
Network enlarging is an effective scheme for generating deep neural networks 
with excellent performance from a smaller baseline. Different from the 
conventional approach that directly enlarge the given network using a unified 
strategy, we present a novel greedy network enlarging algorithm. The entire 
network enlarging task is therefore divided into several iterations for 
searching the best computational allocation in a step-by-step fashion. 
In the enlarging process of the base model, the added MACs will be assigned to 
the most appropriate location. Experimental results on several benchmark models 
and datasets show that the proposed method is able to surpass the original 
unified enlarging scheme and achieves state-of-the-art network performance in 
terms of both network accuracy and computational costs. Beyond allocation of 
MACs in the stage level, more fine grained allocation of MACs are expected.

%\section*{References}

\medskip

\bibliographystyle{plain}
%\bibliography{refer}

%%%%%%%%%%%%%%%%%%%%%%%%%%%%%%%%%%%%%%%%%%%%%%%%%%%%%%%%%%%%

%%%%%%%%%%%%%%%%%%%%%%%%%%%%%%%%%%%%%%%%%%%%%%%%%%%%%%%%%%%%

%\appendix
%
%\section{Appendix}
%
%Optionally include extra information (complete proofs, additional experiments and plots) in the appendix.
%This section will often be part of the supplemental material.

\end{document}